\documentclass[letterpaper, 10 pt, conference]{ieeeconf}  

\IEEEoverridecommandlockouts                              
\usepackage{booktabs}

\usepackage{cite}
\usepackage{amsmath,amssymb,amsfonts}
\usepackage{algorithmic}
\usepackage{graphicx}
\usepackage{textcomp}
\usepackage{xcolor}
\usepackage{subfig}

\usepackage{textcomp}

\usepackage{fancyhdr}
\usepackage[ruled,vlined]{algorithm2e}

\usepackage{amsmath}
\usepackage{fancyhdr,graphicx,amsmath,amssymb}
\usepackage[ruled,vlined]{algorithm2e}
\usepackage{bm}
\usepackage{xcolor}

\usepackage{url}
\usepackage{hyperref}
\hypersetup{
    colorlinks=true,
    linkcolor=blue,
    filecolor=magenta,      
    }

\usepackage{multirow} 
\let\labelindent\relax
\DeclareUnicodeCharacter{2212}{-}

\def\BibTeX{{\rm B\kern-.05em{\sc i\kern-.025em b}\kern-.08em
    T\kern-.1667em\lower.7ex\hbox{E}\kern-.125emX}}

\usepackage{bm} 
\usepackage{color} 

\usepackage{tikz}
\usetikzlibrary{decorations.pathmorphing} 
\usetikzlibrary{matrix} 
\usetikzlibrary{arrows} 
\usetikzlibrary{calc} 
\usepackage{tikzscale}

\tikzstyle{block} = [draw,rectangle,thick,minimum height=2em,minimum width=2em]
\tikzstyle{sum} = [draw,circle,inner sep=0mm,minimum size=2mm]
\tikzstyle{connector} = [->,thick]
\tikzstyle{line} = [thick]
\tikzstyle{branch} = [circle,inner sep=0pt,minimum size=1mm,fill=black,draw=black]
\tikzstyle{guide} = []
\tikzstyle{snakeline} = [connector, decorate, decoration={pre length=0.2cm,
                         post length=0.2cm, snake, amplitude=.4mm,
                         segment length=2mm},thick, magenta, ->]

\usepackage{etoolbox}
\makeatletter
\patchcmd{\@makecaption}
  {\scshape}
  {}
  {}
  {}
\makeatletter
\patchcmd{\@makecaption}
  {\\}
  {.\ }
  {}
  {}
\makeatother
\def\tablename{Table}
\def\tablename{Table}

\usepackage[compact]{titlesec}
\titlespacing{\section}{2pt}{*+1}{*+1}
\titlespacing{\subsection}{2pt}{*+1}{*+1}
\titlespacing{\subsubsection}{2pt}{*+1}{*+1}

\usepackage{enumitem}
\setlist{nolistsep,leftmargin=*}
\setlist{nolistsep}
\usepackage{amsmath}
\usepackage{fancyhdr,graphicx,amsmath,amssymb}
\usepackage[ruled,vlined]{algorithm2e}
\usepackage{bm}
\usepackage{xcolor}
\usepackage[font=footnotesize]{caption}

\begin{document}
\include{pythonlisting}
\title{DAMON: Dynamic Amorphous Obstacle Navigation using \\ Topological
Manifold Learning and Variational Autoencoding
}


\author{
Apan Dastider 
and
Mingjie Lin
}

\maketitle
\begin{abstract}

DAMON leverages manifold learning and variational autoencoding to achieve
obstacle avoidance, allowing for motion planning through adaptive graph
traversal in a pre-learned low-dimensional hierarchically-structured manifold
graph that captures intricate motion dynamics between a robotic arm and its
obstacles. This versatile and reusable approach is applicable to various
collaboration scenarios.

The primary advantage of DAMON is its ability to embed information in a
low-dimensional graph, eliminating the need for repeated computation required
by current sampling-based methods. As a result, it offers faster and more
efficient motion planning with significantly lower computational overhead and
memory footprint. In summary, DAMON is a breakthrough methodology that
addresses the challenge of dynamic obstacle avoidance in robotic systems and
offers a promising solution for safe and efficient human-robot collaboration.

Our approach has been experimentally validated on a 7-DoF robotic manipulator
in both simulation and physical settings. DAMON enables the robot to learn and
generate skills for avoiding previously-unseen obstacles while achieving
predefined objectives. We also optimize DAMON's design parameters and
performance using an analytical framework. Our approach outperforms mainstream
methodologies, including RRT, RRT*, Dynamic RRT*, L2RRT, and MpNet, with 40\%
more trajectory smoothness and over 65\% improved latency performance, on
average.

\end{abstract}


\section{Introduction}

 Ensuring safe collaboration between humans and robots demands that robots be    
equipped to handle uncertainty and partial observability, while executing       
actions in unpredictable and dynamic                                            
environments~\cite{Goodrich2007,Schmitt2019a}. For robotic arms, standard       
algorithms can be effective in motion planning in obstacle-free settings, but   
become more challenging in unstructured environments where the robot's          
workspace is occupied by static and dynamic obstacles~\cite{Falanga2020}.       
Moreover, accomplishing variable or dynamically-changing motion objectives      
significantly raises the difficulty level of active obstacle avoidance.

 Traditionally, obstacle avoidance problems in robotic control relied on              
 dynamical-system-based approaches~\cite{563653,khansari2012dynamical}. These         
 approaches modeled motion dynamics and incorporated probabilistic                    
 methods~\cite{5970128} to handle data variability and model uncertainty.             
 However, recent advancements in Deep Neural Networks (DNN) have prompted             
 researchers to investigate obstacle avoidance from a deep learning perspective,      
 utilizing various reinforcement learning methodologies to capture the intricate      
 motion dynamics between a robotic arm and its obstacles~\cite{Garg2019}.             
                                                                                      
 Despite significant advancements in safe and accurate robotic arm motion             
 planning, several formidable challenges remain. These include: (1) developing a      
 generalized methodology with efficiency and scalability that can quickly adapt       
 to variable targets and robotic arm dynamics, (2) creating a unified framework       
 that can avoid multiple dynamically moving and morphing 3D obstacles without         
 relying on an explicit environment model, and (3) establishing a rigorous            
 mathematical framework to accurately analyze well-defined performance metrics        
 and guide the design tradeoffs of algorithm parameters. Addressing these             
 challenges, among others, is crucial to advancing the field of robotic arm           
 motion planning.                                                             

 \begin{figure}[htbp]                                       
    \centering                                             
    \includegraphics[width=0.8\linewidth]{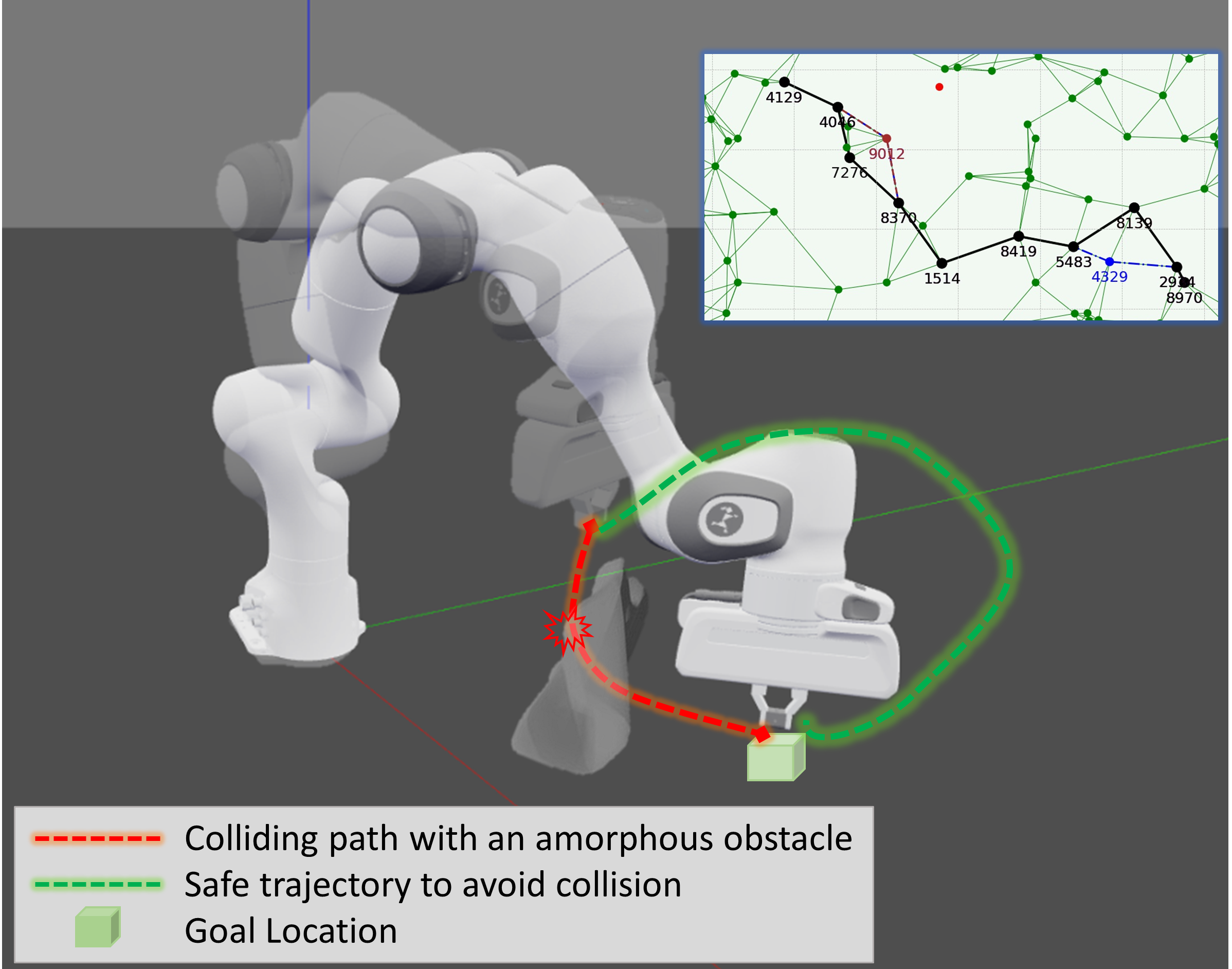}  
    \caption{
        A 7-DoF robotic manipulator needs to reach a target while avoiding a dynamically moving and changing obstacle. From accumulated dataset we learn a variational autoencoder 
        that spans a random topological manifold. The motion control of this manipulator 
        is computed by dynamically traversing a 2D graph generated by manifold learning. 
        }%
    \label{fig:intro}                                   
\end{figure}


\begin{table*}[htbp]
\centering
\begin{tabular}{|ccccccccc|}
\hline
\multicolumn{1}{|c|}{\multirow{2}{*}{Method}}                             & \multicolumn{2}{c|}{\begin{tabular}[c]{@{}c@{}}Planning Data Space\end{tabular}}                      & \multicolumn{2}{c|}{\begin{tabular}[c]{@{}c@{}}Motion Planning Method\end{tabular}}                 & \multicolumn{2}{c|}{\begin{tabular}[c]{@{}c@{}}Obstacle \\ Avoidance\end{tabular}}              & \multicolumn{1}{c|}{\multirow{2}{*}{\begin{tabular}[c]{@{}c@{}}Adaptive\\ Replaning\end{tabular}}} & \multirow{2}{*}{\begin{tabular}[c]{@{}c@{}}Unified Multi-purpose\\ Hierarchical Graph\end{tabular}} \\ 
\cline{2-7}
\multicolumn{1}{|c|}{}                                   & \multicolumn{1}{c|}{Ambient Space}                        & \multicolumn{1}{c|}{Latent Space}                        & \multicolumn{1}{c|}{Graph Search}              & \multicolumn{1}{c|}{Sampling Based}                  & \multicolumn{1}{c|}{Static}                    & \multicolumn{1}{c|}{Dynamic}                   & \multicolumn{1}{c|}{}                                                             &                                                                       \\ \hline
 \hline
\multicolumn{1}{|c|}{DAMON (Ours)}                     & \multicolumn{1}{c|}{}                          & \multicolumn{1}{c|}{\checkmark} & \multicolumn{1}{c|}{\checkmark} & \multicolumn{1}{c|}{}                          & \multicolumn{1}{c|}{\checkmark} & \multicolumn{1}{c|}{\checkmark} & \multicolumn{1}{c|}{\checkmark}                                    & \checkmark                                             \\ \hline
\multicolumn{1}{|c|}{RRT}                                & \multicolumn{1}{c|}{\checkmark} & \multicolumn{1}{c|}{}                          & \multicolumn{1}{c|}{}                          & \multicolumn{1}{c|}{\checkmark} & \multicolumn{1}{c|}{\checkmark} & \multicolumn{1}{c|}{}                          & \multicolumn{1}{c|}{}                                                             &                                                                       \\ \hline
\multicolumn{1}{|c|}{RRT*}         & \multicolumn{1}{c|}{\checkmark} & \multicolumn{1}{c|}{}                          & \multicolumn{1}{c|}{}                          & \multicolumn{1}{c|}{\checkmark} & \multicolumn{1}{c|}{\checkmark} & \multicolumn{1}{c|}{}                          & \multicolumn{1}{c|}{}                                                             &                                                                       \\ \hline
\multicolumn{1}{|c|}{Dynamic RRT*\cite{dynamicrrt}} & \multicolumn{1}{c|}{\checkmark} & \multicolumn{1}{c|}{}                          & \multicolumn{1}{c|}{}                          & \multicolumn{1}{c|}{\checkmark} & \multicolumn{1}{c|}{\checkmark} & \multicolumn{1}{c|}{\checkmark} & \multicolumn{1}{c|}{\checkmark}                                    &                                                                       \\ \hline
\multicolumn{1}{|c|}{L2RRT\cite{ichter2019robot} }                            & \multicolumn{1}{c|}{}                          & \multicolumn{1}{c|}{\checkmark} & \multicolumn{1}{c|}{}                          & \multicolumn{1}{c|}{\checkmark} & \multicolumn{1}{c|}{\checkmark} & \multicolumn{1}{c|}{}                          & \multicolumn{1}{c|}{}                                                             &                                                                       \\ \hline
\multicolumn{1}{|c|}{MpNet\cite{mpnet}}                              & \multicolumn{1}{l|}{}                          & \multicolumn{1}{c|}{\checkmark} & \multicolumn{1}{l|}{}                          & \multicolumn{1}{c|}{\checkmark} & \multicolumn{1}{c|}{\checkmark} & \multicolumn{1}{c|}{\checkmark}                          & \multicolumn{1}{c|}{\checkmark}                                                             & \multicolumn{1}{l|}{}                                                 \\ \hline
\multicolumn{9}{|c|}{\begin{tabular}[c]{@{}c@{}}Notes : (i) Our Proposed Method encompasses both robot pose and obstacle location in one high dimensional vector\\ (ii) One full vectorized geometrical information eliminates the requirement of learning separate collision checker network \end{tabular}}                                                                                                                                                                                                                        \\ \hline
\end{tabular}
\caption{Summarization of various existing methods for Robot Motion Planning against static and dynamic obstacle}
\end{table*}

{\em \bf Related Works}:                                                                           
In recent years, there has been a growing interest in developing smoother                          
robotic motion planning and obstacle avoidance through the formulation of                          
latent space representation and topological manifold learning.                                     
Mohammadi et al. (2021)~\cite{MohammadiHANR21} developed a Riemannian                              
submanifold in $\mathbb{R}^3\times\mathcal{S}^3$ space and used geodesic paths                     
over the learned sub-manifold for robotic motion generation. They also                             
introduced an obstacle avoidance scheme by modifying the ambient metrics in the                    
latent space.                                                                                      
Other works, such as Ichter et al. (2019)~\cite{ichter2019robot}                                   
and MPNet (Motion Planning Networks)~\cite{mpnet}, have                                                  
also introduced sampling-based motion planning and obstacle avoidance through                      
learned latent space networks. Bernstein (2017)~\cite{Bernstein2017ManifoldLI}                     
provided a detailed review of                                                                      
manifold learning algorithms incorporated in recent advancements of machine                        
vision and robotics. Additionally, Khan et al. (2020)~\cite{Khan2020} proposed an optimal          
tracking control and obstacle avoidance solution using recurrent                                   
knowledge-based heuristics and proximal distance measurements between 3D                           
meshes.                                                                                            
Table~\ref{Table2} presents some comparison results between DAMON and 5 recent studies.           
 While L2RRT~\cite{ichter2019robot} and MpNet~\cite{mpnet}                                         
 have introduced manifold representation for learning various                                      
 robotic skills in low-dimensional space, our work DAMON is more scalable and                      
 simpler for real-time shape-changing obstacle avoidance due to the avoidance of                   
 performing expensive sampling-based motion planning. Instead,                                     
 DAMON leverages a densely-connected 2D network of manifold representation of                      
 high-dimensional robotic state space, which allows for efficient and effective                    
 traversal.                                                                                        
 Furthermore, among all these approaches, only DAMON, together with Dynamic RRT*~\cite{dynamicrrt} 
 and MpNet~\cite{mpnet}, can effectively handle dynamic obstacle avoidance.

{\em \bf Statement of Contributions:}                                                       
In this paper, we present DAMON, a novel approach to tackle the challenges of robotic   
arm motion planning. It adopts a topological manifold perspective and                   
adaptive graph traversals to avoid dynamic obstacles, as depicted in                    
Fig.\ref{fig:intro}. Unlike prior works                                                 
\cite{ichter2019robot,563653,5970128,Garg2019} that rely on complex system              
dynamics modeling or multiple-network sample-based computing on latent space,           
DAMON employs a unique manifold learning-based framework that adaptively                
traverses a pre-computed hierarchically-structured graph in a low-dimensional           
latent space.                                                                           
Our specific contributions are as follows:                                              
                                                                                        
(1)                                                                                     
DAMON is superior to previous sampling-based methods due to its universal, versatile, and reusable nature. Once learned for a specific robotic manipulator, it can avoid any number of 3D obstacles with arbitrary and unseen trajectories, making it highly applicable in human-robot collaboration applications. Moreover, DAMON can handle any number of dynamic obstacles with arbitrary shapes once the underlying manifold surface is learned. This makes DAMON more adaptable and robust to real-world scenarios where the environment can change unpredictably.


(2)                                                                                     
DAMON's algorithm creates a hierarchically-structured graph for efficient traversal of the latent space. Its ability to encapsulate intricate information about a robotic arm's interaction with its surroundings in a low-dimensional graph results in faster and more efficient motion planning with lower computational overhead and memory requirements. This scalability and efficiency make it suitable for real-time applications and enable safer human-robot interaction.
                                                                                        
(3)                                                                                     
DAMON uses Gaussian mixture models for statistical learning, allowing for robust performance evaluation and optimal algorithm parameter selection. Its ability to navigate multiple obstacles while achieving multiple objectives makes it versatile and adaptable. DAMON's generalizability means that the trained model can be applied to new environments and obstacles without additional training, saving time and computational resources.


(4)                                                                                 
DAMON's effectiveness was demonstrated in both simulated and real-world scenarios, surpassing state-of-the-art methods in terms of efficiency and effectiveness. It was used to avoid dynamic obstacles with arbitrary shapes using a 7-DoF robotic manipulator. The approach learned and replicated complex robot skills, and could handle new obstacles without requiring additional learning efforts.


\section{Problem Formulation: Motion Planning for Dynamic Obstacle Avoidance}

 The motion control of a robotic manipulator involves calculating a joint-space     
 trajectory that guides the end-effector to a desired position. The forward         
 kinematic mapping of a $k$-DOF robotic manipulator in an $n$-dimensional task      
 space is a surjective function of the joint-space coordinates,                     
 $                                                                                  
 \mathbf{x}(t) = f(\mathbf{\theta}(t))                                              
 $,                                                                                 
 where                                                                              
 $\mathbf{x}(t) \in \mathbb{R}^n$ and $\mathbf{\theta}(t) \in \mathbb{R}^k$ are     
 the task-space and joint-space coordinates, respectively. This nonlinear           
 vector-valued function can be easily formulated using the mechanical design and    
 Denavit-Hartenberg parameters for a given manipulator. However, for most           
 applications, computing the inverse mapping from the task space to the joint       
 space is more important, as it allows us to specify the desired end-effector       
 position. Similarly, we can define an inverse kinematics model as                  
 $                                                                                  
 \mathbf{\theta(t)}= f^{-1}(\mathbf{x}(t))$,                                        
 where $f^{-1}(\cdot)$ is the inverse kinematic mapping.                            
                                                                                    
 Our goal is to solve the inverse kinematics equation for the joint-space           
 coordinates that guide the end-effector to a target position. However, solving     
 this equation alone does not guarantee that the calculated trajectory will         
 avoid collisions with obstacles. To address this issue, we formulate the           
 problem of obstacle avoidance as the problem of maximizing the minimum distance    
 between the links of the manipulator and the obstacle.

Let $\mathcal{S}\subseteq\mathbb{R}^{s^n}$ and $\mathcal{U}\subseteq\mathbb{R}^k$      
defines the state space and control input of a robotic manipulator    
system such that $\mathcal{S}_{r}(t), \mathcal{S}_{o}(t) \in \mathcal{S}$ where   
$\mathcal{S}_{r}(t)$ defines the robot's state condition at timestep, $t$ and     
$\mathcal{S}_{o}(t)$ carries all information about independent obstacles in the   
environment. 
The robotic system evolves through time by following defined         
discrete-time dynamics of robotic system,                                         
\begin{equation}                                                                  
    \label{dyn}                                                                   
    \mathcal{S}(t+1) = \mathcal{F}_{\mathcal{S}}(\mathcal{S}(t), \mathcal{U}(t))       
\end{equation}                                                                    
where, $\mathcal{F}_{\mathcal{S}}$ defines the functional for discrete-time       
system evolution. Now, let $\mathcal{S}_{free}\in\mathcal{S}$ and                 
$\mathcal{S}_{coll}\in\mathcal{S}$ defines the free state space and colliding     
state space of the robot, such that $\mathcal{S}_{free} = \mathcal{S}             
\symbol{92} \mathcal{S}_{coll}$. Now for any initial state                        
$\mathcal{S}_{init}\in\mathcal{S}_{free}$ and a defined goal state                
$\mathcal{S}_{goal}\in\mathcal{S}_{free}$, we aimed at finding a continuous       
trajectory, $\mathcal{T}:=(\mathcal{S}_{init},u_{init},                      
\cdots\mathcal{S}_{i},u_{i},\cdots\mathcal{S}_{goal},u_{goal})$, such   
that the continuous curve between adjacent state space in the planned             
trajectory remains \textit{collision free} i.e, $(\mathcal{S}_{i},                
\mathcal{S}_{i+1})\cap\mathcal{S}_{coll}=\emptyset$. Our algorithm ensured that   
all trajectory path safely ends at pre-defined goal state $\mathcal{S}_{goal}$.


\section{Proposed Methodology}

\begin{figure}[htbp]                                       
    \centering                                             
    \includegraphics[width=\linewidth]{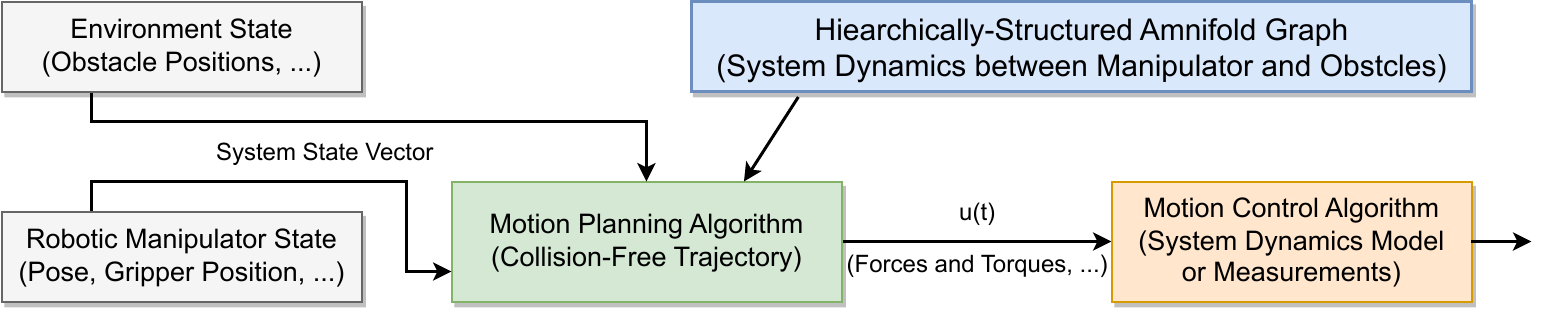}  
    \caption{
        Algorithmic block digram of DAMON methodology.
        }%
    \label{fig:algo-idea}                                   
\end{figure}                                               

Fig.~\ref{fig:algo-idea} depicts DAMON, consisting mainly of two algorithm          
modules: the motion control algorithm (MCA) and the motion planning algorithm       
(MPA).                                                                              
The MPA module computes an optimized trajectory that satisfies all spatial and      
temporal requirements, while the MCA generates control signals that regulate        
the position, velocity, and acceleration of the robotic manipulator's               
actuators, allowing the robot manipulator to track the computed trajectory.         
Although DAMON adopts the conventional PID control for MCA, it focuses on           
developing an innovative motion planning algorithm that efficiently avoids          
dynamic obstacles while reaching its objective state.                               
The key idea behind DAMON is to adaptively traverse a hierarchically-structured     
manifold graph that captures the intricate dynamics of the entire system,           
including a whole-body robotic arm and a point obstacle located anywhere in the     
workspace. By doing so, DAMON efficiently avoids dynamic obstacles while            
achieving its objective state.                                                      
                                                                                    
DAMON addresses the primary challenge of computational inefficiency and             
intractability in motion planning for real-world robotic systems. Such systems      
have complex system dynamics and high-dimensional state-space, making               
conventional algorithms unsuitable for dynamic environments.                        
Even sampling-based motion planning methods can become inefficient when the         
state-space evolves randomly after the planner has completed planning.              
Moreover, sampling-based dynamic replanning algorithms, such as                     
\cite{dynamicrrt,mpnet}, consume a significant amount of runtime to re-sample a     
new path when dynamic obstacles block the initial path. This inefficiency is        
further compounded by the fact that the computational load increases as the         
number of obstacles and dimensions in the workspace grows.

\begin{figure}[htbp]                                       
    \centering                                             
    \includegraphics[width=\linewidth]{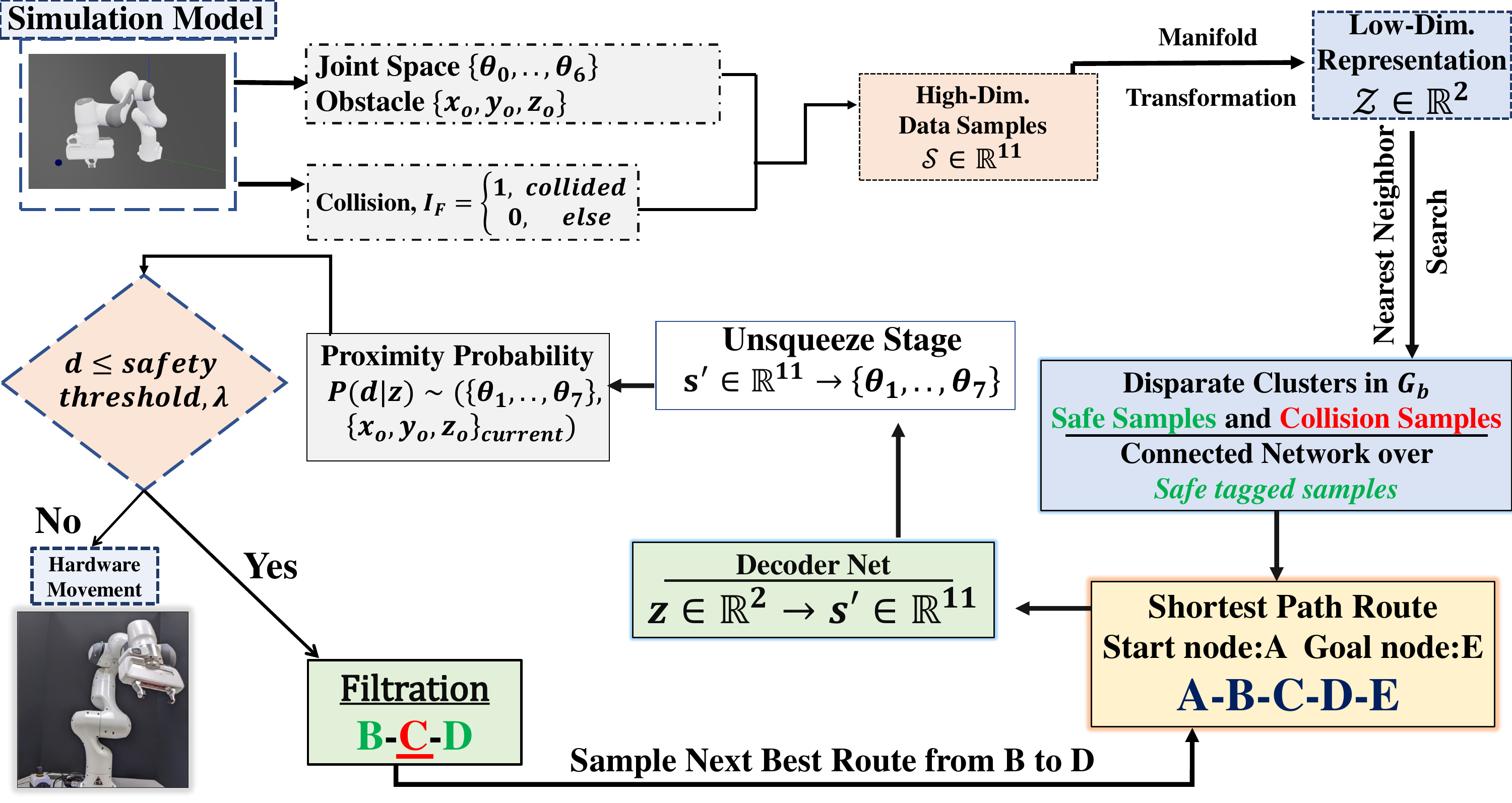}  
    \caption{
        Flow diagram of algorithmic stages of DAMON.
        }%
    \label{fig:method}                                   
\end{figure}                                               

In the following, we provide more information on the three major algorithm 
modules that constitute DAMON.                                             

\subsection*{\bf Algorithm Module 1: Topological Manifold Learning}

Manifold learning is a mathematical framework used to investigate the
geometrical structure of datasets in high-dimensional spaces. 
In this paper, we
consider a high-dimensional space defined by $[\theta_0, \cdots, \theta_6;
[x,y,z]_o; \mathbb{I}_F]$, where $\theta_i$ determines the full joint-space
pose of a robotic arm and $[x, y, z]_o$ defines the location of a point
obstacle. $\mathbb{I}_F$ is a boolean collision flag.
Our key insight of DAMON is that the geometrical structure of this
high-dimensional space, combining both the pose of robotic manipulator and the
3D location of obstacle, can encapsulate all the intricate information about
how a robotic arm interacts with its surroundings, especially a point obstacle
at an arbitrary location,  in a low-dimensional latent-space graph. This
approach circumvents the need for repeated computation required by obstacle
detection and motion replanning.  
Furthermore, to handle 3D geometrical obstacles, we
consider the closest point on the 3D mesh of the object to 3D collision meshes
of the robot's pose as the point obstacle coordinate showed in Figure \ref{fig:point_obs}.
%

\begin{figure}[htbp]                                       
    \centering                                             
    \includegraphics[width=1.0\linewidth]{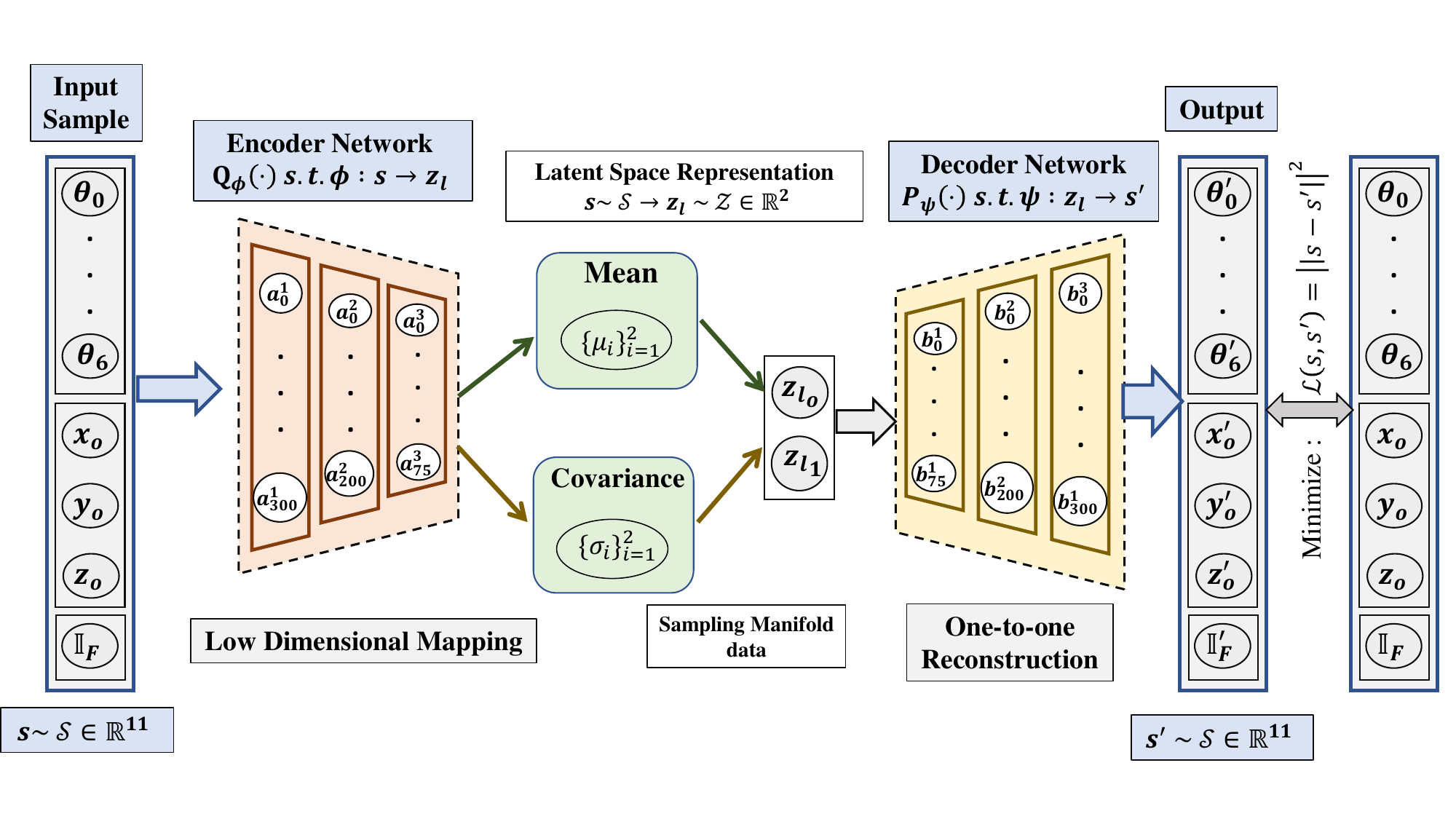}  
    \caption{
        Schematic of Variational autoencoder used for Manifold Learning.
        }%
    \label{fig:autoencoder}                                   
\end{figure}                                               

\begin{figure}[htbp]                                       
    \centering                                             
    \includegraphics[width=\linewidth]{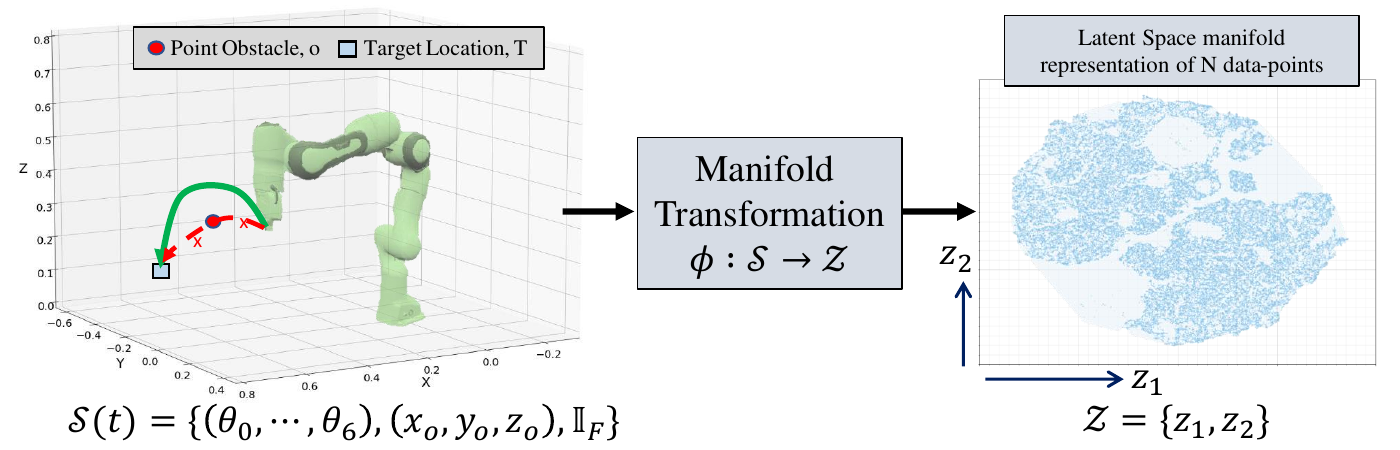}  
    \caption{The high-dimensional dataspace transforms to Latent Manifold representation to facilitate efficient and scalable motion planning
        }%
    \label{fig:method1}                                   
\end{figure}

%

To leverage the embedding power of our learned low-dimensional manifold for
robotic motion planning, we need to seamlessly transform between the
high-dimensional robotic space, $\mathcal{S}$, and the low-dimensional manifold
space, $\mathcal{Z}$. To achieve this, we utilized a variational autoencoder
(VAE)~\cite{kingma2014autoencoding} for both latent space learning and decoding
to the high-dimensional state space.
Specifically, as depicted in Fig.~\ref{fig:autoencoder}, our VAE is implemented
as a feedforward, non-recurrent neural network that employs an input layer and
an output layer connected by multiple hidden layers. The output layer has the
same number of nodes (neurons) as the input layer.
The purpose of our VAE is to reconstruct its inputs, minimizing the difference
between the input and the output, rather than predicting a target value
$\mathcal{S'}$ given inputs $\mathcal{S}$. This allows us to capture the
essential information about the high-dimensional robotic space in a
lower-dimensional manifold, enabling us to perform efficient motion planning
and obstacle avoidance in real-time.
The implementation of VAE in DAMON largely adopted from~\cite{mpnet} and is trained
through backpropagation of the error.
Conceptually, the feature space $z_l$ of our autoencoder is the low-dimension
manifold representations produced by manifold learning, therefore having lower
dimensionality than the input space $\mathcal {S}$, which, in DAMON, is the
pose of our 7-DoF robotic manipulator with object position and collision flag.
As such, the feature matrix $Q_\phi (s)$ after manifold learning can be
regarded as a compressed representation of the input $s$.


\subsection*{\bf Algorithm Module 2: Hierarchical Graph Construction and Graph-Traversal-Based Motion Planning}



To achieve robustness and adaptability in motion planning for real-world
scenarios where the environment can change unpredictably, DAMON employs two key
techniques. Firstly, we learn the latent space manifold representation
$\mathcal{Z} \in \mathbb{R}^{z^n}$, where $z^n \ll s^n$, through a mapping
function $\phi:\mathcal{S}\rightarrow\mathcal{Z}$. Secondly, instead of relying
on a sampling-based tree expansion, we construct a fully connected graph
$G=(V,E)$, where $V_i:=z_i\in\mathcal{Z}$ and $E$ is weighted by nearest joint
space control inputs for smoother trajectory execution. We utilize $G$ for path
traversal from any $V_{start}$ to $V_{goal}$. Furthermore, the dense
connectivity of our graph allows the planner to re-route from any node $V_{i}$
without significant runtime loss, even if the old path becomes convoluted due
to a potential collision with a moving obstacle.
These two techniques ensure that DAMON can efficiently handle changing
environments while maintaining adaptability and robustness. The latent space
manifold representation reduces the dimensionality of the state-space, making
it easier to handle and learn. The fully connected graph enables us to utilize
the learned representation for efficient path traversal and re-routing, even in
dynamic environments with moving obstacles.

Although graph-based robotic motion planning is a well-established
technique~\cite{8968151}, it is commonly used for mobile robots operating in complex
environments\cite{KupervasserDMRobot, chenDMRobot}. 
Our DAMON methodology has two distinctions from these graph-based motion planning studies.
Firstly, DAMON exploits a
two-layered hierarchical graph that encodes the complex system dynamics between
a robotic manipulator and its closest obstacle through a topological manifold
space.
Secondly, DAMON performs adaptive graph traversal for effective motion planning. 

\begin{figure}[htbp]                                       
    \centering                                             
    \includegraphics[width=1.0\linewidth]{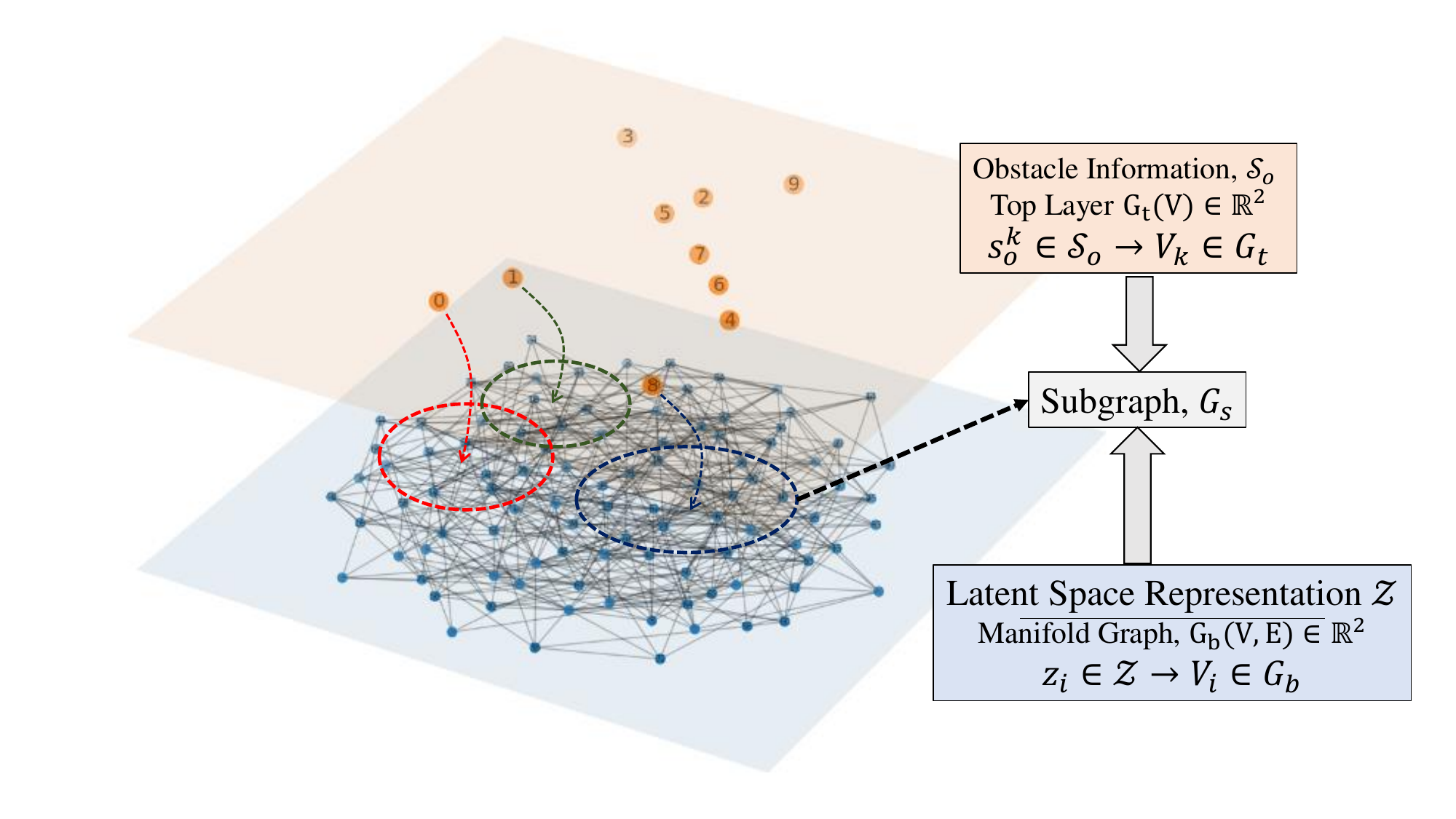}  
    \caption{
        Hierarchical Graph Structure.
        }%
    \label{fig:hie_graph}                                   
\end{figure}

DAMON simplifies its data structure and facilitates graph traversal by
leveraging a hierarchically-structured low-dimensional manifold space
$\mathcal{Z} \in \mathbb{R}^2$. In this manifold graph depicted in Figure~\ref{fig:hie_graph}, 
the bottom layer $G_b$
contains all the manifold points $\{z_i\}_{i=1}^n \in \mathcal{Z}$ learned from
VAE for the stored dataset, with each $z_i$ corresponding to a vertex $V_i$ in
the graph layer $G_b$. The top layer $G_t$ considers the point obstacle
$\mathcal{S}_{o}^{k}$ information as each node $V_k$, where $k$ is the number of
obstacles considered while accumulating the dataset.

To further simplify the structure, the manifold graph is partitioned into
multiple subgraphs, each containing all vertices with the same obstacle 3D
locations. Conceptually, a hyperedge between the two layers connects the
vertices $\{V_i\}_{i=1}^m \in G_b$ associated with respective $V_k \in G_t$.
Furthermore, the vertices in each subgraph are labeled as either "collision" or
"collision-free/safe" samples. During robotic arm motion planning, we use
Dijkstra's algorithm to perform shortest-distance routing, considering only the
green vertices ("collision-free/safe") for safer robot movement.

\begin{figure}[htbp]                                       
    \centering                                             
    \includegraphics[width=1.0\linewidth]{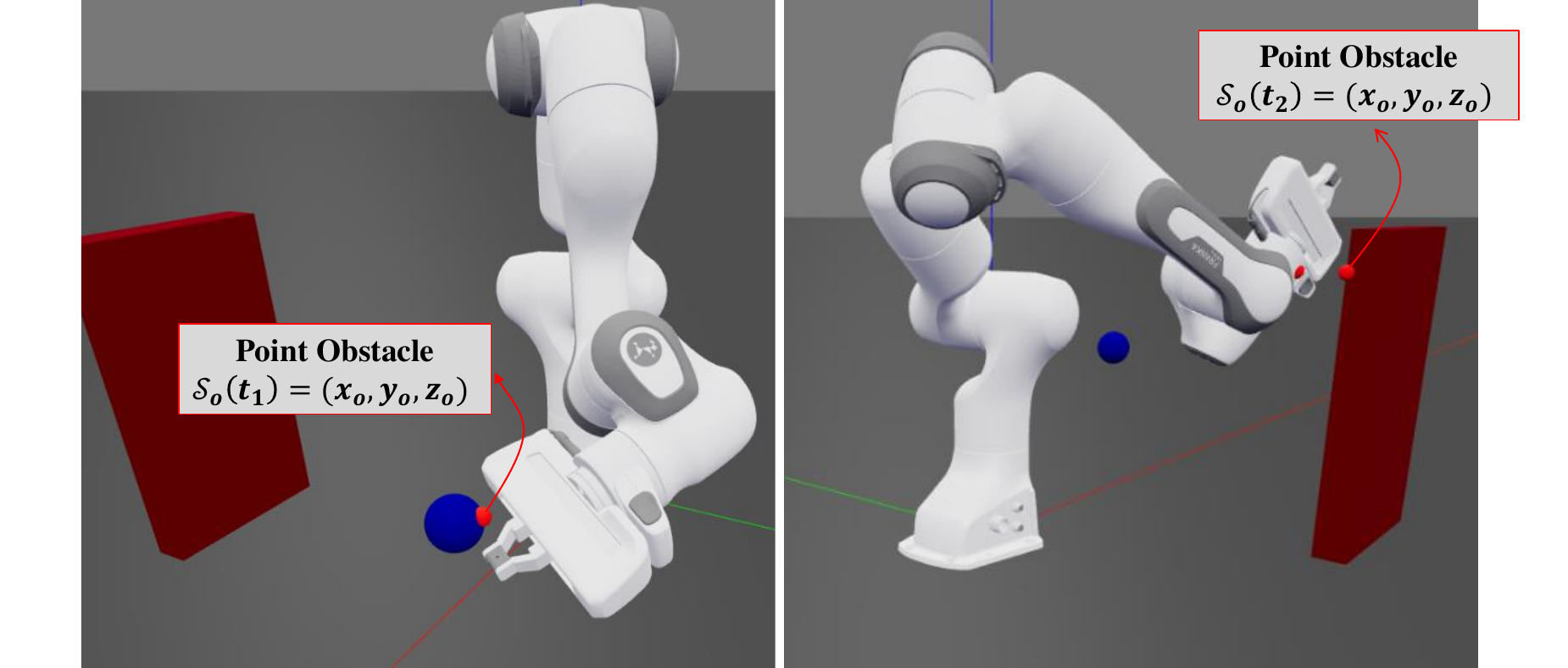}  
    \caption{
        Here, $t_1<t_2$. The point location information gets updated as the robot moves close to cuboid obstacle after avoiding the spherical obstacle.
        }%
    \label{fig:point_obs}                                   
\end{figure}

The 2-layer hierarchically-structured manifold graph reduces computational
overhead as DAMON only needs to traverse the subgraph $G_s$ in $G_b$ that is
closest to the current obstacle point location. We use a KD-Tree to quickly
query the node $V_k$ that is closest to the current point obstacle location, 
as illustrated in Fig.~\ref{fig:point_obs}.
When multiple dynamic obstacles are present, DAMON hops among different
subgraphs according to the one closest to the robotic manipulator.

Our implementation manipulates graph-like objects solely via predefined API
methods and not by acting directly on the data structure. We utilize the
well-known "dict-of-dicts" structure as the main data structure, which allows
fast addition, deletion, and lookup of nodes and neighbors in large graphs. Our
software code is based largely on the open-source NetworkX
package\cite{SciPyProceedings_11}. We omit the standard implementation details
for the sake of brevity.

\subsection*{\bf Algorithm Module 3: Theoretical Framework for Performance Analysis and Algorithm Parameter Optimization}
\label{sec:mod3}

\begin{figure}[htbp]                                       
    \centering                                             
    \includegraphics[width=\linewidth]{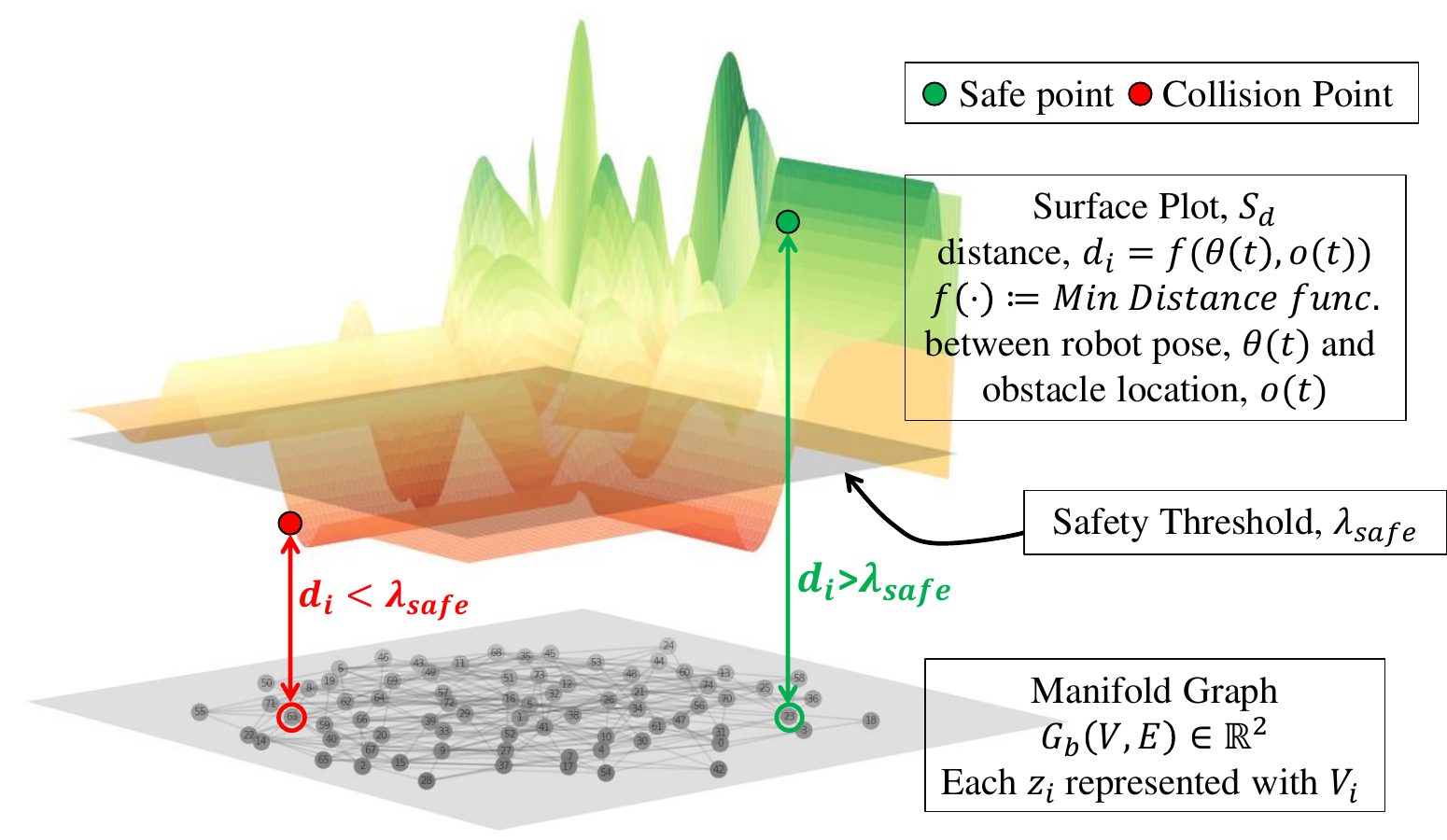}  
    \caption{Each latent space representation $z_i$ is associated with the minimum distance $d_i$ between the robot's collision meshes and nearest obstacle point. We encapsulate this data to perform parameter optimization for DAMON.
        }%
    \label{fig:surface}                                   
\end{figure}
When optimizing the algorithm of DAMON, we need to address two key algorithm     
questions.  Firstly, {\em how to accurately estimate the delay performance of    
DAMON for a given task and its setting?} Secondly, {\em for a given delay        
performance requirement, what should be the ideal density of our constructed     
manifold graph?}  To satisfactorily answer both of these questions, DAMON        
develops an analytical framework to compute the average rerouting probability    
and derive the relationship between the total runtime of each robotic            
experiment and the density of our low-dimensional manifold graph.  To this end,  
DAMON utilizes the Gaussian mixture regression (GMR)~\cite{NIPS1993_f2201f51}    
to estimate the conditional probability distribution of $d_i$ given a set of     
input latent space variables, i.e., its manifold coordinates $\mathbf{z}_i$ of   
vertex $V_i$, as shown in Figure~\ref{fig:surface}.                              

 DAMON assumes that our manifold graph $\mathcal{G}$ consists of $N$ vertices, each of which   
 is denoted by $V_i$ defined by low-dimensional coordinates $\mathbf{z}_i$ and augmented with  
 $d_i \in \mathtt{R}$. Abstractly,  $\mathcal{G}$ represents a multivariate functional surface  
 depicted in Fig.~\ref{fig:surface}.                                                           
 With the GMR method,                                                                          
 DAMON assumes that the target variable $d_i$ obeys a mixture of                                
 Gaussian distributions, where the parameters of each Gaussian component are dependent on      
 the input variables $\mathbf{z}_i$, where $i=0., 1., \cdots, {N-1}$.                          
 During the training phase, DAMOM learns a $K$-component Gaussian mixture model                
 $                                                                                             
 p(\mathbf{z}, \mathbf{d})=\sum_{k=1}^{K}                                                      
 \pi_{k} \mathcal{N}_{k}\left(\mathbf{z}, d \mid                                               
 \boldsymbol{\mu}_{\mathbf{z} d_{k}},                                                          
 \boldsymbol{\Sigma}_{\mathbf{z} d_k}\right)                                                     
 $                                                                                             
 through Expectation-Maximization (EM) training~\cite{NIPS1993_f2201f51}, where                
 \(\mathcal{N}_{k}\left(\mathbf{z},                                                            
 d \mid \boldsymbol{\mu}_{\mathbf{z} d_{k}},                                                   
 \boldsymbol{\Sigma}_{\mathbf{z} d_{k}}\right)\)                                               
 are Gaussian distributions with mean                                                          
 \(\boldsymbol{\mu}_{\mathbf{z} d_{k}}\)                                                       
 and covariance \(\boldsymbol{\Sigma}_{\mathbf{z} d_{k}}, K\)                                  
 is the number of Gaussians, and \(\pi_{k} \in[0,1]\) are priors that sum up to one.           
 After the Gaussian mixture model is successfully trained,                                     
 DAMON performs a regression to predict distributions of variables $d_i$                       
 by computing the conditional distribution $p(d_i \mid \mathbf{z}_i)$.                         
  The conditional distribution of each individual Gaussian is                               
 $                                                                                           
 \mathcal{N}\left(\mathbf{z}, {d} \mid \boldsymbol{\mu}_{\mathbf{z} d},                      
 \boldsymbol{\Sigma}_{\mathbf{z} d}\right) $,                                                
 where                                                                                       
 $                                                                                           
 \boldsymbol{\mu}_{\mathbf{z} d}=\left(\begin{array}{l}                                      
     \boldsymbol{\mu}_{\mathbf{z}} \\                                                        
 {\mu}_{d}                                                                                   
 \end{array}\right)$,                                                                        
 $\boldsymbol{\Sigma}_{\mathbf{z} d}=\left(\begin{array}{cc}                                 
     \boldsymbol{\Sigma}_{\mathbf{z} \mathbf{z}} & \boldsymbol{\Sigma}_{\mathbf{z} d} \\     
     \boldsymbol{\Sigma}_{d \mathbf{z}} & {\Sigma}_{d d}                                     
 \end{array}\right)                                                                          
 $,                                                                                          
 $                                                                                           
 \mu_{d \mid \mathbf{z}}=\mu_{d}+\boldsymbol{\Sigma}_{d \mathbf{z}}                          
 \boldsymbol{\Sigma}_{\mathbf{z} \mathbf{z}}^{-1}                                            
 \left(\mathbf{z}-\boldsymbol{\mu}_{\mathbf{z}}\right)                                       
 $,                                                                                          
 and                                                                                         
 $                                                                                           
 \Sigma_{d \mid \mathbf{z}}=\Sigma_{d d}-\boldsymbol{\Sigma}_{d \mathbf{z}}                  
 \boldsymbol{\Sigma}_{\mathbf{z} \mathbf{z}}^{-1} \boldsymbol{\Sigma}_{\mathbf{z} d}         
 $.                                                                                          
 {DAMON} can now compute the conditional distribution of each individual Gaussian and their priors 
according to                                                                                      
$                                                                                                 
\pi_{d \mid \mathbf{z}_{k}}=                                                                      
\frac{\mathcal{N}_{k}\left(\mathbf{z} \mid \boldsymbol{\mu}_{\mathbf{z} k},                       
\boldsymbol{\Sigma}_{\mathbf{z} k}\right)}{\sum_{l=1}^{K}                                         
\mathcal{N}_{l}\left(\mathbf{z} \mid \boldsymbol{\mu}_{\mathbf{z} l},                             
\boldsymbol{\Sigma}_{\mathbf{z} l}\right)}                                                        
$                                                                                                 
to obtain the conditional distribution                                                            
$                                                                                                 
p(d \mid \mathbf{z})=\sum_{k=1}^{K} \pi_{d \mid \mathbf{z}_{k}}                                   
\mathcal{N}_{k}\left(d \mid \boldsymbol{\mu}_{d \mid \mathbf{z}_{k}},                             
\boldsymbol{\Sigma}_{d \mid \mathbf{z}_{k}}\right)                                                
$.                                                                                                
Now, given any location $\mathbf{z}$ on the manifold graph, we can compute its probability of     
collision as                                                                                      
\begin{equation}                                                                                  
\begin{aligned}                                                                                   
    & p({\rm collision} | \mathbf{z}) =                                                           
p(0 \le d \le \lambda | \mathbf{z})                                                                
    = \int_{0}^{\lambda} p( y| \mathbf{z}) dy \\                                                   
    & =  \int_{0}^{\lambda} \left[ \sum_{k=1}^{K} \pi_{{y} \mid \mathbf{z}_{k}} \mathcal{N}_{k}    
    \left({y} \mid \boldsymbol{\mu}_{{y} \mid \mathbf{z}_{k}}, \boldsymbol{\Sigma}_{{y}           
    \mid \mathbf{z}_{k}}\right) \right] dy \\                                                     
    & =  \int_{0}^{\lambda} \left[ \sum_{k=1}^{K}                                                  
    {                                                                                             
\frac{\mathcal{N}_{k}\left(\mathbf{z} \mid \boldsymbol{\mu}_{\mathbf{z}_k},                       
\boldsymbol{\Sigma}_{\mathbf{z}_k}\right)                                                         
\cdot                                                                                             
    \mathcal{N}_{k}                                                                               
    \left({y} \mid \boldsymbol{\mu}_{{y} \mid \mathbf{z}_{k}}, \boldsymbol{\Sigma}_{{y}           
    \mid \mathbf{z}_{k}}\right)                                                                   
    }{\sum_{l=1}^{K}                                                                              
\mathcal{N}_{l}\left(\mathbf{z} \mid \boldsymbol{\mu}_{\mathbf{z}_l},                             
\boldsymbol{\Sigma}_{\mathbf{z}_l}\right)}                                                        
    }                                                                                             
    \right] dy,                                                                                   
\end{aligned}                                                                                     
    \label{eqn:4}                                                                                 
\end{equation}                                                                                    
where $\lambda$ is a user-defined parameter that defines the minimum distance required to be       
collision-free between obstacle and the robotic manipulator.

Let one complete trajectory in DAMON contain $K$ vertices $V_i$,                               
$i= 0, 1,\cdots, K-1$ on the manifold surface graph.                                           
During each step from $V_i \rightarrow V_{i+1}$,                                               
the robotic manipulator will take                                                              
$t_i = \frac{t_{i, {p}}}{1 - p({\rm collision} \mid V_i)} + t_{i, {r}}$                        
in total run-time, where                                                                       
$t_{i, {p}}$ and $t_{i, {r}}$ denote the motion planning time for each step and                
the runtime for the robotic manipulator moving from $V_i$ to $V_j$,                            
$i= 0, 1,\cdots, K-1$.                                                                         
Note that the multiplying factor of $1/(1 - p({\rm collision} \mid V_i))$ accounts for the     
rerouting time when traversing $V_i$ causes collision.                                         
Therefore, the total runtime for each task trajectory will follow                              
                                                                                               
\begin{equation}                                                                               
T= \sum_{i=0}^{K-1}  t_i =                                                                     
\sum_{i=0}^{K-1} \frac{t_{i, {p}}}{1 - p({\rm collision} \mid V_i)} + t_{i, {r}}.              
    \label{eqn:5}                                                                              
\end{equation}                                                                                 
                                                                                               
\subsubsection{Estimating the delay performance of DAMON}                                      
                                                                                               
DAMON uses the standard EM learning algorithm to extract a Gaussian Mixture                    
Model $\mathcal{M} (\mathcal{G})$ with statistical accuracy. With $\mathcal{M}                 
(\mathcal{G})$ in hand, we can easily calculate the probability of collision at                
any location $V_i$ on the manifold, as indicated by Equation~\ref{eqn:4}. This                 
approach provides an effective means of estimating the total runtime for each                  
trajectory task, as shown in Equation~\ref{eqn:5}.                                             
It's worth noting that we have made two simplifying assumptions. First, we                     
assume that each routing step on the manifold graph is probabilistically                       
independent during each computed trajectory. Second, we assume that the robotic                
manipulator moves at a constant average rate. These assumptions were made for                  
the sake of brevity in modelling. However, it is possible to achieve more                      
complex modelling to account for variable velocity and acceleration.                           

\subsubsection{Determining the optimal manifold point density}                     
                                       
Intuitively, as the point density of the manifold graph increases, the             
rerouting probability $p({\rm collision} \mid V_i)$ due to data sparsity error     
decreases. However, the graph traversing time and the motion planning time in      
each step $t_{i,p}$ are likely to increase because a denser manifold graph         
requires more computations. Additionally, as the rerouting probability at each     
manifold vertex decreases, the average number of steps decreases. Hence, there     
is an intriguing and complicated relationship between the number of manifold       
points collected and the average runtime for a typical motion trajectory. Our      
experimental results in Fig.~\ref{fig:rprob} have validated this analytical finding.        
Equations~\ref{eqn:4} and~\ref{eqn:5} together provide an effective way of         
determining the optimal number of manifold points we need to collect, given a      
well-defined performance requirement in terms of the total runtime for each        
task.                                                                              
\begin{figure*}[htbp]
    \includegraphics[width=1\textwidth]{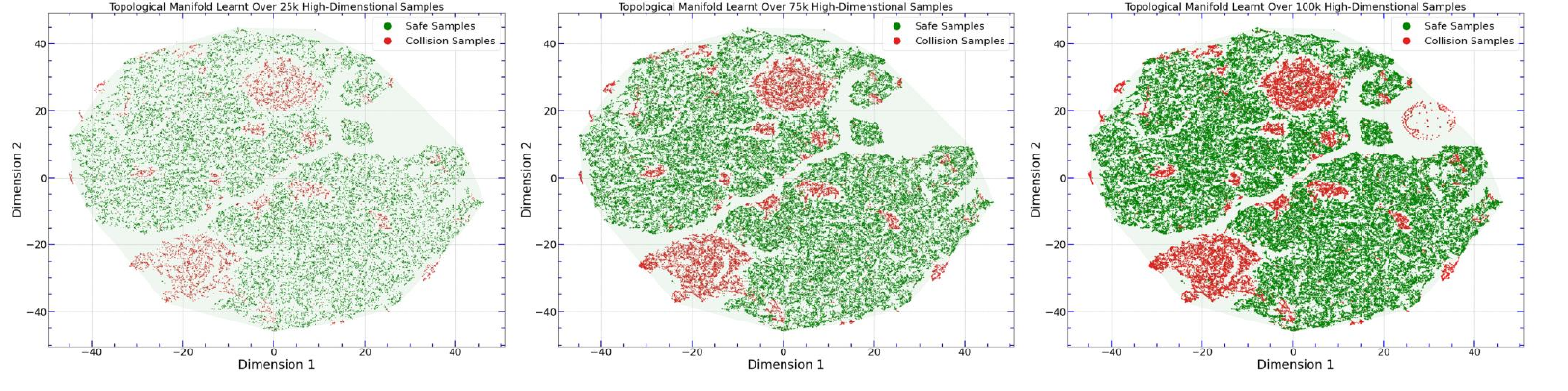}
    \caption{Structure of Topological Manifold with varying number of input samples. (a) for 10k samples, (b) for 25k samples, (c) for 100k samples}
    \label{fig:manifold}
\end{figure*}                               

\begin{figure*}[htbp]
    \includegraphics[width=1\textwidth]{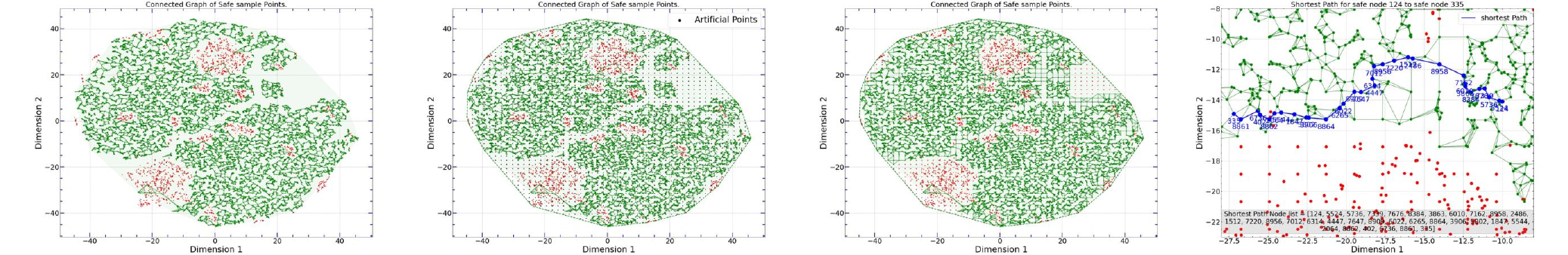}
    \caption{(a) Connected Network over 10k Manifold Points (To have a clear visual, we plotted the network with 10k points. All planning were completed with optimal samples), (b)-(c) Uniformly Sampled Points for a densely Connected Network, (d) Shortest Path Routing by Dijkstra's algorithm }
    \label{fig:shortest}
\end{figure*}


\section{System Overview}
\label{sec:system}
\subsection{Experimental Platform and Simulation Setup}

In our hardware demonstrations, we utilized a 7-DoF Franka Emika Panda robot arm mounted on a table-top workspace, as depicted in Fig.\ref{fig:hardware}. Our adaptive trajectory planning algorithm was implemented in Python and ran on a Lambda QUAD GPU workstation equipped with an Intel Core-i9-9820X processor. To track dynamic obstacles and extract depth information, we used an Intel RealSense Depth Camera D435i. By applying the conventional hand-to-eye calibration mathematical formulation, we transported the $[x,y,z]_o$ information from the camera reference frame to the robot coordinate system. Our algorithm utilized this feedback from depth sensors to avoid obstacles in real-world scenarios by computing the distance between the 3D geometry of the obstacle and the collision meshes of robot links using the Gilbert–Johnson–Keerthi distance (GJK)\cite{Khan2020} algorithm to safely route to the target location. To validate our approach, we replicated the exact model of the Franka Emika Panda Arm in Robotics Toolbox (RTB) for Python. We encapsulated the entire framework inside the Robot Operating System (ROS) ecosystem, utilizing libfranka and Franka ROS, to establish low-latency and low-noise communication protocols for data processing and parallel execution among simulation environments and real hardware.



\subsection{Experimental Procedures and Learning Variables}

We collected high-dimensional data samples from the robot's workspace and transformed them into a latent space representation for creating the topological manifold. Each data sample contained 11 numerical variables, including 7 joint angles, the position of a point obstacle in 3D space, and a binary collision flag. We accumulated a dataset of 1MM samples by randomly manipulating the arm in the simulated environment for a large number of epochs. Our VAE architecture, implemented on PyTorch~\cite{Pytorch}, had a decoder and encoder network with three layers containing (300, 200, 75) neurons and an encoded latent space representation in $\mathbb{R}^2$. We focused on collecting adequate collision samples to ensure a safer elementary trajectory. When the algorithm and manifold learning converged, we conducted real-time and scalable test experiments against varying occlusions created by geometrically different 3D objects in the real hardware setup.

\section{Results Analysis}

Our experiments seek to investigate the following: 






\begin{enumerate}

\item Can DAMON learn manifold representation 
    $\mathcal{Z}\in\mathbb{R}^2$ space differentiating the
        non-colliding and colliding samples with hierarchical structure for faster routing? 
        
    \item  Can we preserve the best performance model with optimal and computationally efficient sample density for efficient graph routing while reducing the rerouting probability?

\item Can DAMON concurrently track the unseen perturbations created by geometrically varying obstacles and dynamically adjust the trajectory to reach the goal
        location through graph routing? 

\end{enumerate}


\subsection{Latent-Space Manifold Graph and Optimum Routing}


 Each input sample in DAMON's high-dimensional space comprises the joint angle          
 vector for the robotic manipulator in operation, the 3D coordinate location of         
 the closest point on the obstacle mesh, and a collision flag $I_F$.                    
 Our Variational Autoencoder (VAE)                                                      
 learns a topological manifold representation $\mathcal{Z}\in\mathbb{R}^2$ from         
 this high-dimensional space. As illustrated in Fig.~\ref{fig:manifold},                
 the latent space manifold representation exhibits visually contrasting clusters        
 based on the binary level of the collision flag.  Next, we construct a                 
 connected graph, $G_b$, over the $\mathbb{R}^2$ space that contains only the           
 collision-free samples. By applying the unsupervised $K$-Nearest Neighbor              
 algorithm~\cite{knn} to locate the nearest family of vertices in the collision-free    
 manifold space, we create a connected graph that accelerates any                       
 shortest path routing algorithm to traverse on a transformed low-dimensional           
 manifold graph. We further divide the large graph, $G_b$, into several                 
 subgraphs, $G_s$, where each $G_s$ is associated with its own obstacle                 
 information embedded in the top layer $G_t$ of the hierarchical graph                  
 structure.  As depicted in Fig.\ref{fig:shortest}(a), disconnected sub-networks        
 are automatically generated inside the safe manifold while creating connected          
 graphs, as the nearest neighboring algorithm only helps to create a connected          
 network at the closest distance. This sub-network generation violates the              
 notion of a complete path routing from any random node to another random node          
 on the fully connected network. To ensure dense connectivity, we artificially          
 sample points in the same $\mathbb{R}^2$ space as a grid mesh over the learned         
 $\mathbb{R}^2$ space. By adjusting the level of sparsity in creating the grid          
 of artificial points, we can easily sample denser artificial points to create a        
 densely connected network. With the decoder part of our trained VAE, we can            
 label these artificial points as safe/colliding sample points, as shown in             
 Fig.\ref{fig:shortest}(c). Furthermore, with decoded high-dimensional points,          
 we can label these artificial points to their nearest obstacle node of $G_t$           
 and include them in the associated subgraph networks.                                  

\begin{table*}[htbp]
\centering
\begin{tabular}{|ccccccccccccc|}
\hline
\multicolumn{1}{|c|}{\multirow{3}{*}{Methods}}           & \multicolumn{5}{c|}{Environment Settings}                                                                                                                                               & \multicolumn{5}{c|}{Environment Settings}                                                                                                                                                                    & \multicolumn{2}{c|}{Obstacle Condition}                             \\ \cline{2-13} 
\multicolumn{1}{|c|}{}                                   & \multicolumn{5}{c|}{Total Runtime, T (s)}                                                                                                                                           & \multicolumn{5}{c|}{Smoothness  Comparison}                                                                                                                                                                  & \multicolumn{2}{c|}{Success Ratio (\%SR)}                           \\ \cline{2-13} 
\multicolumn{1}{|c|}{}                                   & \multicolumn{1}{c|}{A}             & \multicolumn{1}{c|}{B}             & \multicolumn{1}{c|}{C}             & \multicolumn{1}{c|}{D}             & \multicolumn{1}{c|}{E}              & \multicolumn{1}{c|}{A}                  & \multicolumn{1}{c|}{B}                  & \multicolumn{1}{c|}{C}                 & \multicolumn{1}{c|}{D}                 & \multicolumn{1}{c|}{E}                 & \multicolumn{1}{c|}{\;\;\;\;\;\;\;\;\;\;Static\;\;\;\;\;\;\;\;\;\;}                & Dynamic                \\ \hline \hline
\multicolumn{1}{|c|}{DAMON}                              & \multicolumn{1}{c|}{3.2}           & \multicolumn{1}{c|}{4.7}           & \multicolumn{1}{c|}{5.1}           & \multicolumn{1}{c|}{6.5}           & \multicolumn{1}{c|}{7.3}            & \multicolumn{1}{c|}{0.31}               & \multicolumn{1}{c|}{0.35}               & \multicolumn{1}{c|}{0.41}              & \multicolumn{1}{c|}{0.57}              & \multicolumn{1}{c|}{0.51}              & \multicolumn{1}{c|}{98.6}                  & 97.8                   \\ \hline
\multicolumn{1}{|c|}{RRT}                                & \multicolumn{1}{c|}{42.5}          & \multicolumn{1}{c|}{39.2}          & \multicolumn{1}{c|}{46.3}          & \multicolumn{1}{c|}{-}             & \multicolumn{1}{c|}{-}              & \multicolumn{1}{c|}{0.25}               & \multicolumn{1}{c|}{0.25}               & \multicolumn{1}{c|}{0.25}              & \multicolumn{1}{c|}{-}                 & \multicolumn{1}{c|}{-}                 & \multicolumn{1}{c|}{78.4}                  & -                      \\ \hline
\multicolumn{1}{|c|}{RRT*}                               & \multicolumn{1}{c|}{22.6}          & \multicolumn{1}{c|}{20.4}          & \multicolumn{1}{c|}{29.6}          & \multicolumn{1}{c|}{-}             & \multicolumn{1}{c|}{-}              & \multicolumn{1}{c|}{0.61}               & \multicolumn{1}{c|}{0.81}               & \multicolumn{1}{c|}{0.55}              & \multicolumn{1}{c|}{-}                 & \multicolumn{1}{c|}{-}                 & \multicolumn{1}{c|}{83.2}                  & -                      \\ \hline
\multicolumn{1}{|c|}{Dynamic-RRT*}                       & \multicolumn{1}{c|}{15.1}          & \multicolumn{1}{c|}{13.4}          & \multicolumn{1}{c|}{15.9}          & \multicolumn{1}{c|}{37.2}          & \multicolumn{1}{c|}{45.2}           & \multicolumn{1}{c|}{0.55}               & \multicolumn{1}{c|}{0.67}               & \multicolumn{1}{c|}{0.41}              & \multicolumn{1}{c|}{0.91}              & \multicolumn{1}{c|}{0.84}              & \multicolumn{1}{c|}{86.2}                  & 76.3                   \\ \hline
\multicolumn{1}{|c|}{L2RRT}                              & \multicolumn{1}{c|}{7.9}           & \multicolumn{1}{c|}{6.2}           & \multicolumn{1}{c|}{8.3}           & \multicolumn{1}{c|}{-}             & \multicolumn{1}{c|}{-}              & \multicolumn{1}{c|}{0.35}               & \multicolumn{1}{c|}{0.38}               & \multicolumn{1}{c|}{0.39}              & \multicolumn{1}{c|}{-}                 & \multicolumn{1}{c|}{-}                 & \multicolumn{1}{c|}{90.2}                  & -                      \\ \hline
\multicolumn{1}{|c|}{MpNet}                              & \multicolumn{1}{c|}{5.6}           & \multicolumn{1}{c|}{6.1}           & \multicolumn{1}{c|}{7.2}           & \multicolumn{1}{c|}{10.3}          & \multicolumn{1}{c|}{12.85}          & \multicolumn{1}{c|}{0.42}               & \multicolumn{1}{c|}{0.53}               & \multicolumn{1}{c|}{0.64}              & \multicolumn{1}{c|}{0.88}              & \multicolumn{1}{c|}{0.95}              & \multicolumn{1}{c|}{91.3}                  & 82.3                   \\ \hline\hline
\multicolumn{13}{|c|}{Environment Settings Details:}\\ \hline
\multicolumn{6}{|l|}{\begin{tabular}[c]{@{}l@{}} \scriptsize A : Static Obstacle \#1 Shape : Cuboidal Block\\ \scriptsize  B : Static Obstacle \#1 Shape : Spherical Block\\ \scriptsize  C : Static Obstacle \#2 Shape : Cuboidal,  Spherical\end{tabular}} & \multicolumn{7}{l|}{\begin{tabular}[c]{@{}l@{}} \scriptsize D : Static \#1 and Dynamic \#1 Shape : Cuboidal (\#1, Static), Spherical (\#1, Dynamic)\\  \scriptsize E : Static \#2 and Dynamic \#1 Shape : Cuboidal (\#2, Static), Spherical (\#1, Dynamic)\end{tabular}} \\ \hline
\end{tabular}
\caption{Comparison Results between DAMON and Baseline Approaches for varying environments}
\label{Table2}
\end{table*}

\subsection{Performance Analysis and Comparison to Baselines}

We evaluate the performance of DAMON against classical navigation approaches and state-of-the-art 
latent-space sampling-based algorithms with the following metrics:
\begin{itemize}
    \item \textit{Total Runtime, $T$}: Total Runtime is determined by calculating total time to compute the initial path, $t_{i,p}$; time to maneuver by the physical robot joints, $t_{i,r}$; time to perform replanning if the initial path gets occluded with dynamic obstacle, $t_{p(collision|V_i)}$;
        $T = f(t_{i,p} ,t_{p(collision|V_i)}) + t_{i,r}$
    
    \item \textit{Success Ratio, $SR$}: A planned trajectory is successful if the robotic arm reaches the target location without colliding with any moving obstacles.
    
    \item \textit{Trajectory Smoothness:} Smoothness 
    will be measured by summing up the joint space configuration change in consecutive via-point pair 
    along the planned path, i.e $\sum_{i=0}^{N-1}||\Theta_i - \Theta_{i+1}||_2$, where N is the total number of movement stages.
    
\end{itemize}

Table~\ref{Table2} showcases the performance of DAMON in comparison to other main methods. To ensure thoroughness, we have tested DAMON in 5 different settings that involve both static and dynamic obstacles to verify its generalization capability. It is worth noting that RRT, RRT*, and L2RRT do not explicitly support dynamic obstacle replanning; hence, their results against dynamic environments are unavailable.
Regarding total runtime, DAMON outperforms other approaches by 2-3 times on average across all cases. However, when it comes to trajectory and movement smoothness, the comparison results are mixed. In general, DAMON is smoother than RRT* Dynamic RRT*, and MpNet, but less smooth than RRT, and about the same as L2RRT. We believe that this discrepancy is primarily due to the sampling-intensive nature of computing versus relatively more efficient graph traversal.
Finally, in terms of navigation success rate, DAMON clearly demonstrates its edge over all other methods. We attribute this mostly to the synergistic interplay between topological manifold learning and adaptive graph traversal.

\begin{figure}[htbp]
    \centering
    \includegraphics[width=\linewidth]{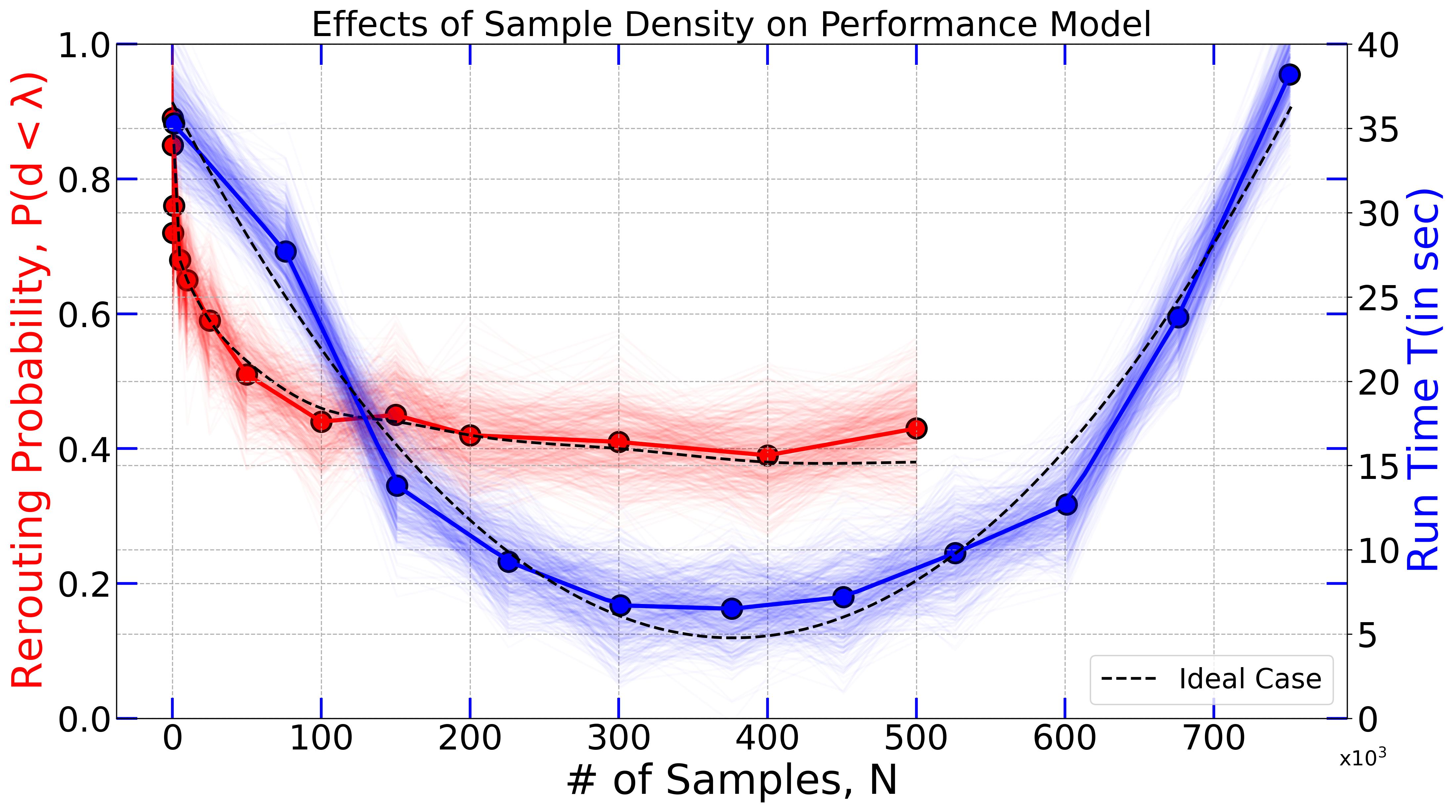}
    \caption{Analysis of how rerouting probability evolves as we changed number of nodes in the graph structure (In Red). Total Run Time Performance Analysis for increasing number of Nodes in the Graph Structure(In Blue).}%
    \label{fig:rprob}
\end{figure}
Furthermore, Fig.~\ref{fig:rprob} plots how the number of samples $N$ 
affects the total runtime $T$ and the aggregated rerouting probabily $P$ in DAMON 
across 12 different obstacle scenarios.
When tested against with the analytic performance framework in Section~\ref{sec:mod3}, 
our experimental results have shown a close match.
For both cases, larger number of sampling points to form our manifold graph will in general make the 
graph traversal more successful and rerouting less needed. However, 
denser manifold graph also requires more computation, thus long per-stage motion planning time.
As such, there is an intriguing and complicated interplay between the density of manifold graph 
and performance metrics.
To balance this trade-off between readily available safe path for rerouting and initial trajectory over optimal number of nodes $n^*$, we found the optimal number of nodes $n^*$ which produced the lowest average $T$ over set of different test-cases. Lastly, Figure \ref{fig:rprob} showed how the rerouting probability plateaued to a fixed range as we consistently increased the number of samples while computing rerouting probability. This convergence of rerouting probability pointed out that we located $n^*$ which allowed required amount of rerouting capability without increasing additional computational overhead for our graph-centric navigation planning.       


\begin{figure}[htbp]
    \centering
    \includegraphics[width=\linewidth]{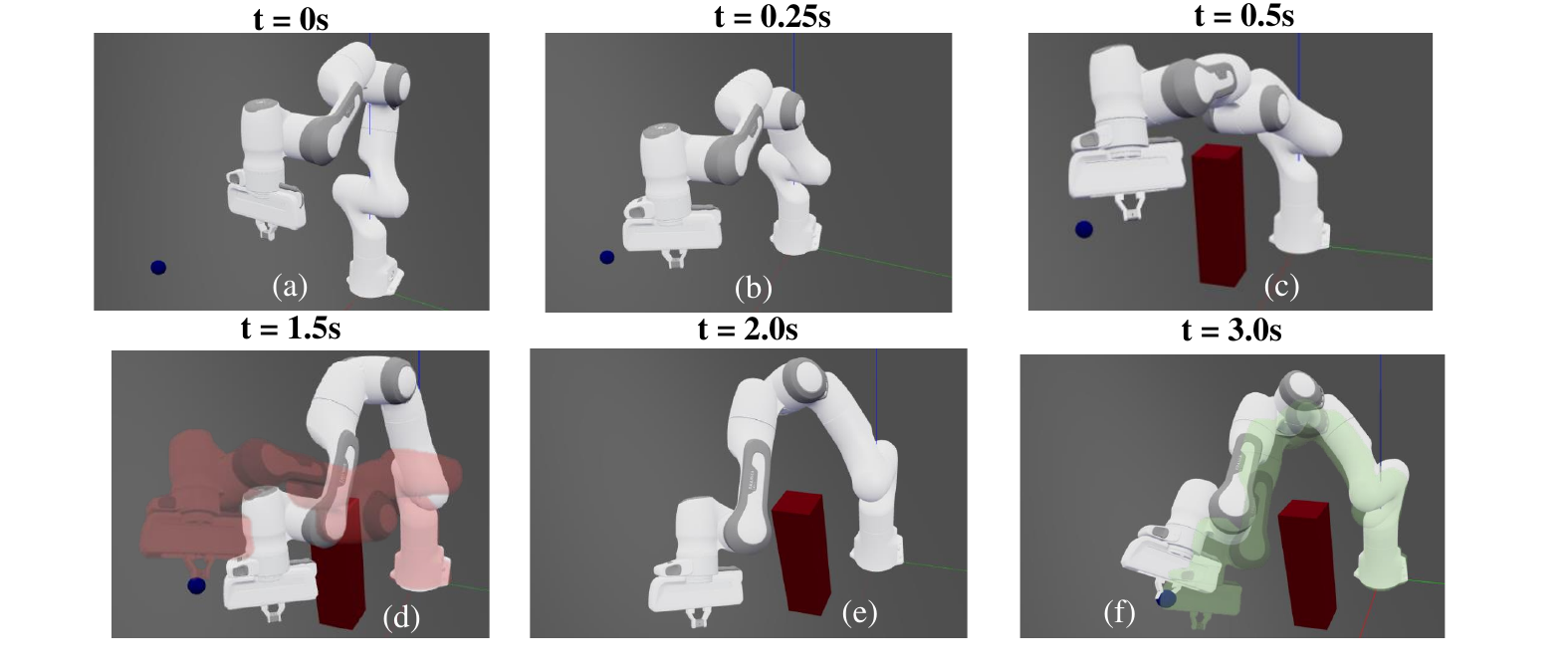}
    \caption{(a) Initial pose, (b) Robot follows its initial path, (c) Obstacle appears at different time-step and robot stops it previous motion, (d) Robot revises its trajectory to avoid possible collision, in red shaded -- collision when the trajectory planning \textit{non-adaptive}, (e) (f) Safely reach target}%
    \label{fig:sim1}
\end{figure}

\subsection{Adaptive Navigation with Dynamic Obstacle Avoidance}


DAMON achieves dynamic obstacle avoidance on a 2-D
manifold graph, by adapting the trajectory in real-time to reach the goal
position as quickly as possible while avoiding moving obstructions. The system
achieves this by receiving feedback based on distance measurements and adapting
its path using a nearest neighbor network.  It is worth noting that each
manifold vertex in DAMON contains not only the pose state of the robotic arm
and a specific point obstacle, but also the precomputed distance between the
obstacle and the robotic manipulator. This precomputed information and 
the topology of manifold graph is crucial
because it embeds the intricate dynamics between the robotic arm and the
obstacle, effectively minimizing the computing effort required for on-the-fly
motion planning.

\begin{figure*}[htbp]
    \centering
    \includegraphics[width=\linewidth]{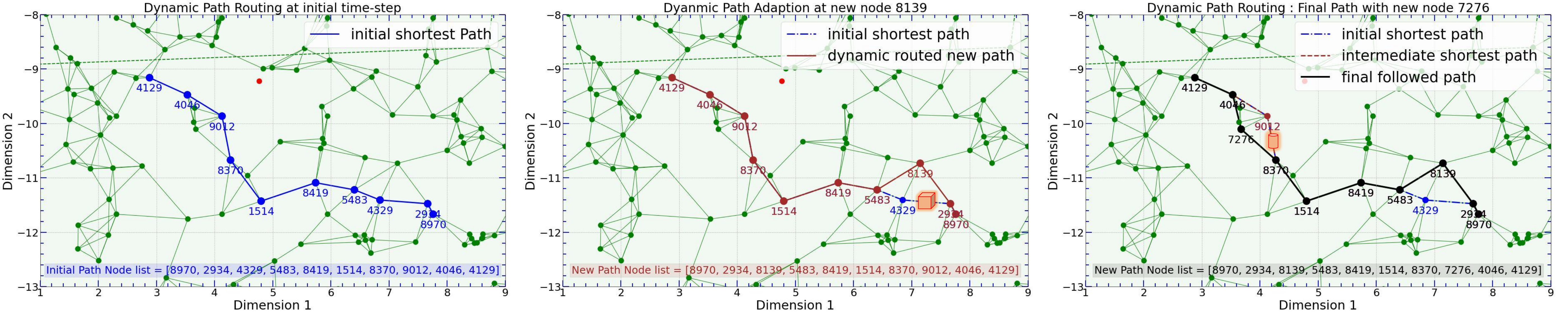}
    \caption{Addressing the presence of any amorphous obstacles and avoiding collision through sampling new path from the connected networks}%
    \label{fig:amo}
\end{figure*}
We demonstrate the effectiveness of our proposed method in achieving dynamic
obstacle avoidance with Fig.~\ref{fig:sim1}.  Initially, the robotic arm
attempted to reach the goal position by following its initial computed
trajectory. However, at a later time step, as shown in Fig.~\ref{fig:sim1}(c),
a dynamically placed obstacle blocked the robot's motion. DAMON effectively
guided the robotic arm to re-route through a new trajectory, which successfully
dodges the obstacle and move around the red block, thus avoiding a possible
collision and successfully reaching the target position.  With extensive
experimentations, as listed in Table~\ref{Table2}, DAMON has shown to be
capable of efficiently adapting the robot manipulator's trajectory in
real-time, enabling it to avoid moving obstructions and reach its destination
safely.

\begin{figure*}[htbp]
    \includegraphics[width=1\textwidth]{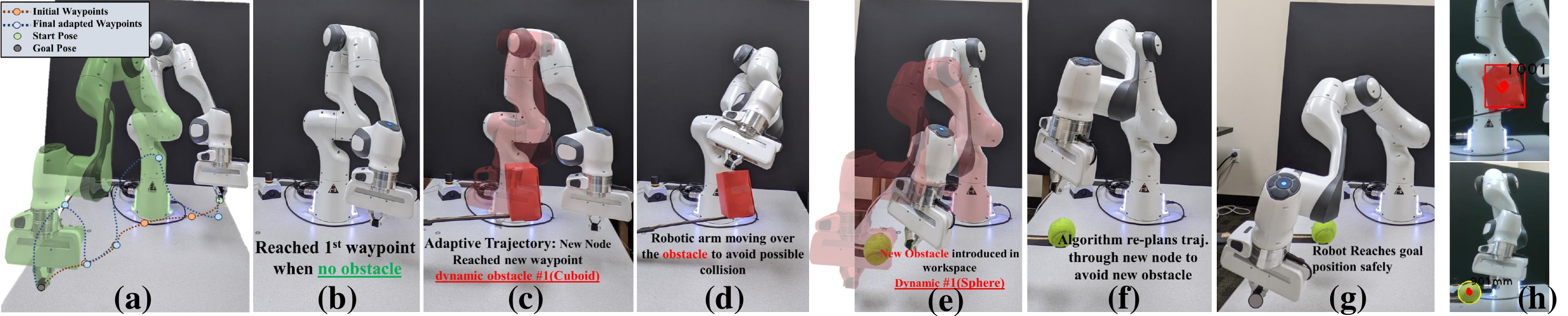}
    \caption{(a) Comparison between initial waypoints and final waypoints, (c) Presence of obstacle compels the robot to update previous path, (d) Routing over the obstacle to avoid collision, (e) Reached next waypoint while again perturbed by different shaped-spherical obstacle, (f) algorithm enables the arm to dodge the collision, (g) Robot arm reaches its goal position successfully, (h)  Object Detection and Depth Information}
    \label{fig:hardware}
\end{figure*}
Since DAMON leverages only point obstacle information for replanning, our
proposed approach can be easily extended for avoiding probable collision with
any shaped obstacles. In Fig.~\ref{fig:amo}, we attempted to graphically convey
an idea on how DAMON successfully performed replanning when the robot's initial
motion was convoluted with presence of different shaped obstacles. For
instance, the routing performed successfully even when the 3D shape of
obstacles varied between a cuboid mesh and cylindrical mesh.  Inside the RTB
simulated environment, integrated high functionality API can be effortlessly
used to track the proximal distance among 3D meshes. This API has been modified
with the GJK algorithm~\cite{Khan2020} for calculating the proximity among 3D
meshes in real-world scenarios. This distance metric produced a feedback to the
controller when the robot's mechanical structure reaches very close to an
obstacle. When there is chance of probable collision i.e $P(d \le \lambda)$, the
robot controller is triggered to halt its current trajectory motion and revise
its intermediate node connectivity to reach the preset goal position. On
hardware setup, we localized obstacles in consecutive frames by applying color
filtering through HSV color model and tracked the spatial positions of random
obstacles through depth information extracted from Realsense depth camera.
After successful extraction of depth values, we transported the 3D coordinate
value of closest point on the object mesh to the learning controller and the
controller evaluated the probability of collision based on our GMM model. If
the condition is suddenly violated, the algorithm replans for a new shortest
and safest path to reach the goal position. In Fig.~\ref{fig:hardware}, we have
added snapshots from real hardware operation.  In red shaded figure, we also
depicted the probable collision would occur without the presence of dynamic
adaption on real time. Here, we validated our experiments with varying size and
different geometric objects. For both obstacles, our adaptive motion planning
approach successfully avoided the dynamic obstacles which appear at different
time-step of robot operation.       

\section{Conclusion}

Dynamically avoiding 3D obstacles with unpredictable trajectories for a given
robotic manipulator can be effectively addressed by combining three theoretical
algorithm modules - topological manifold learning, variational autoencoding,
and adaptive graph traversing. By reducing dynamic obstacle avoidance to a
basic point obstacle avoidance problem supported by a strong theoretical
foundation, DAMON demonstrates versatility and generalizability in tackling
arbitrary numbers of obstacles.


%


\bibliographystyle{./bibliography/IEEEtran}
\bibliography{bib}

\end{document}